# TrialMatchAI: An End-to-End AI-powered Clinical Trial Recommendation System to Streamline Patient-to-Trial Matching

**Authors:**

Majd Abdallah[1,2], Sigve Nakken[3,4,5], Mariska Bierkens[6], Johanna Galvis[1,2], Alexis Groppi[1,2], Slim Karkar[1,2], Lana Meiqari[6], Maria Alexandra Rujano[9], Steve Canham[9], Rodrigo Dienstmann[7,8], Remond Fijneman[6], Eivind Hovig[3,5], Gerrit Meijer[6], Macha Nikolski[1,2]

**Affiliations:**
[1] University of Bordeaux, CNRS, IBGC UMR 5095, 146 Rue Léo Saignat, 33000 Bordeaux, France.
[2] University of Bordeaux, Bordeaux Bioinformatics Center, 146 Rue Léo Saignat, 33000 Bordeaux, France.
[3] Department of Tumor Biology, Institute of Cancer Research, The Norwegian Radium Hospital, Oslo University Hospital, Oslo, Norway.
[4] Centre for Cancer Cell Reprogramming, Institute of Clinical Medicine, Faculty of Medicine, University of Oslo, 0379 Oslo, Norway.
[5] Department of Informatics, University of Oslo, 0316 Oslo, Norway.
[6] Department of Pathology, The Netherlands Cancer Institute, Amsterdam, The Netherlands.
[7] Oncology Data Science (ODysSey) Group, Vall d'Hebron Institute of Oncology (VHIO), Barcelona, Spain.
[8] University of Vic - Central University of Catalonia, Barcelona, Spain
[9] European Clinical Research Infrastructure Network (ECRIN), Boulevard Saint Jacques 30, 75014, Paris, France.

**Corresponding authors:** Majd Abdallah (abdallahmajd7@gmail.com), Macha Nikolski (macha.nikolski@u-bordeaux.fr)

## Abstract

Patient recruitment remains a major bottleneck in clinical trials, calling for scalable and automated solutions. We present TrialMatchAI, an AI-powered recommendation system that automates patient-to-trial matching by processing heterogeneous clinical data, including structured records and unstructured physician notes. Built on fine-tuned, open-source large language models (LLMs) within a retrieval-augmented generation framework, TrialMatchAI ensures transparency and reproducibility and maintains a lightweight deployment footprint suitable for clinical environments. The system normalizes biomedical entities, retrieves relevant trials using a hybrid search strategy combining lexical and semantic similarity, re-ranks results, and performs criterion-level eligibility assessments using medical Chain-of-Thought reasoning. This pipeline delivers explainable outputs with traceable decision rationales. In real-world validation, 92% of oncology patients had at least one relevant trial retrieved within the top 20 recommendations. Evaluation across synthetic and real clinical datasets confirmed state-of-the-art performance, with expert assessment validating over 90% accuracy in criterion-level eligibility classification—particularly excelling in biomarker-driven matches. Designed for modularity and privacy, TrialMatchAI supports Phenopackets-standardized data, enables secure local deployment, and allows seamless replacement of LLM components as more advanced models emerge. By enhancing efficiency, interpretability and offering lightweight, open-source deployment, TrialMatchAI provides a scalable solution for AI-driven clinical trial matching in precision medicine.

# 1. Introduction

The advancement of personalized medicine relies heavily on clinical trials, which rigorously evaluate the efficacy and safety of novel therapeutic strategies and validate actionable biomarkers [1, 2]. Yet, a critical bottleneck persists: the timely and efficient recruitment of eligible patients. This challenge not only delays access to potentially life-saving treatments, but also leads to significant resource inefficiencies, hindering the translation of research into clinical practice [3]. Furthermore, only a small fraction of eligible patients are enrolled in clinical trials, despite the potential benefits. Addressing this problem requires scalable and efficient patient-trial matching solutions to accelerate trial completion and ensure that research findings are approved for clinical practice in a timely manner [4].

Traditionally, patient-trial matching relies on a labor-intensive manual review of patient records and trial eligibility criteria, often performed by multidisciplinary teams such as Molecular Tumor Boards (MTBs) in oncology [5, 6]. This process is inefficient, unscalable, and prone to missed enrollment opportunities, particularly in high-stakes areas like pediatric oncology, where timely trial access is critical [7]. The unstructured nature and considerable volume of trial eligibility criteria further exacerbate these challenges, making manual approaches increasingly unsustainable and underscoring the urgent need for efficient automation [8–11].

Early automation efforts relied on rule-based logic and probabilistic systems, which, while effective for structured scenarios, struggled with the semantic diversity and contextual nuances of clinical text [12–17]. Deep learning approaches have addressed some of these limitations by improving feature extraction and handling sequential dependencies inherent in clinical texts [13]. However, these models often rely on large, well-annotated datasets, which are scarce in the biomedical domain, limiting their scalability and generalizability [18, 19].

Recent advances in natural language processing (NLP), particularly through Large Language Models (LLMs), have opened new avenues for processing and interpreting complex clinical texts. Pre-trained LLMs excel at capturing long-range dependencies and contextual relationships, allowing for the generation of clinically meaningful embeddings [20–25]. LLMs have even shown impressive ability to match patients to clinical trials without being explicitly trained for this task, achieving results comparable to human experts [10, 20, 26, 27]. However, most existing LLM-based trial matching systems, such as TrialGPT [10], rely heavily on proprietary, API-driven models, creating barriers related to cost, accessibility, reproducibility, and, critically, patient data privacy and regulatory compliance (e.g., GDPR, HIPAA [28]). The dependence on black-box, closed-source solutions also hinders transparency and prevents other researchers from building upon or adapting these models for specific clinical needs.

To address these limitations, we introduce TrialMatchAI, a fully open-source, locally deployable clinical trial recommendation system designed to ensure transparency, security, and unrestricted research accessibility while eliminating reliance on proprietary LLMs. This ensures compliance with regulatory frameworks, promotes interpretability in patient-trial matching decisions, and supports continuous model improvement as new data becomes available. Unlike previous systems that require external API-based models, TrialMatchAI enables complete control over the trial-matching process, ensuring compliance with stringent data privacy regulations.

Built on a Retrieval-Augmented Generation (RAG) framework and fine-tuned for medical reasoning, TrialMatchAI combines contextual understanding and explainability in eligibility classification, overcoming the limitations of existing rules-based and machine learning-based trial matching solutions. Importantly, the system is highly-relevant for oncology, where patient-trial matching is often complicated by the need to integrate and interpret diverse clinical, molecular, and genetic data, including biomarker expression profiles and genomic mutations. This is made possible by a dedicated LLM-driven data processing module that extracts, normalizes, and standardizes biomolecular information, such as genes, proteins, and mutations from both patients records and clinical trial eligibility criteria. Furthermore, TrialMatchAI is designed for interoperability, seamlessly integrating with electronic health record (EHR) systems via standardized data exchange formats such as Phenopackets [29]. Finally, the modular design enables easy adaptation to new LLM architectures and domain-specific fine-tuning, ensuring that the system remains up to date with the latest biomedical advancements.

The framework was evaluated using synthetic datasets, comprising the commonly-used benchmarks from the TREC clinical trials challenge (years 2021 and 2022) [30, 31], and a custom-built "Ideal Candidates" dataset. Moreover, we leveraged a real-world cohort of 52 cancer patients from the Netherlands Cancer Institute (NKI) [32]. Expert assessments were also conducted on 1050 patient-criterion pairs, validating the system's high accuracy in criterion-level eligibility classification. Our results on the synthetic TREC datasets demonstrate that TrialMatchAI retrieves >90% of relevant trials within the top-ranked 3% of a broad search space containing >26,000 trials in each dataset. Results on the real cancer patients dataset further support these results, showing that 92% of patients had a relevant trial retrieved within the top 20 recommendations. Finally, expert evaluation of criterion-level matching on both synthetic and real patient datasets shows that TrialMatchAI's medical Chain-of-Thought (CoT) model [33] achieves over 90% accuracy in criterion-level classification and explanation generation. These results show that TrialMatchAI, outperforming existing AI-based tools that rely on significantly larger, proprietary GPT (Generative Pre-trained Transformer) models.

In summary, TrialMatchAI is a scalable, privacy-preserving patient-trial matching system, optimized for oncology. It integrates multi-modal data, supports local deployment, and leverages fine-tuned LLMs within a RAG framework to improve recruitment, especially for biomarker-driven trials. This paper details its architecture, evaluation, and performance in real-world and synthetic benchmark settings.

## 2. Results

### 2.1 TrialMatchAI: A Modular AI System for Patient-Trial Matching

Clinical trial eligibility criteria are complex and often formatted as unstructured free texts, requiring a system that can efficiently process heterogeneous and semantically-rich patient data. TrialMatchAI addresses this challenge by leveraging a suite of open-source and fine-tuned, Large Language Models (LLMs) within a Retrieval-Augmented Generation (RAG) framework [34]. RAG enables the system to anchor its reasoning in retrieved trial information, improving both accuracy and transparency in patient-trial matching. TrialMatchAI's modular design allows for efficient text parsing, embedding, classification, and re-ranking, ensuring accurate and context-aware matching. Moreover, the flexible design of TrialMatchAI enables seamless integration of new models and optimization strategies, ensuring adaptability to evolving clinical needs.

**System Overview and Core Components**

TrialMatchAI is designed to handle heterogeneous clinical data, including structured attributes (e.g., age, sex, primary diagnosis, lab results) and unstructured sources (e.g., prior treatments, physician notes, pathology reports). To ensure interoperability with hospital systems and external research databases, the system adopts the Phenopackets exchange format [29], enabling standardized representation of patient data across diverse sources. Moreover, this format, with its capacity to integrate both structured and unstructured data, enables effective LLM-based semantic processing of clinical narratives.

To identify relevant clinical trials for a patient, TrialMatchAI employs a hybrid search and retrieval strategy, combining lexical search [35, 36] for keyword-based matching with vector search [37] using dense embeddings to capture deeper semantic relationships. Following initial retrieval, LLM-based re-ranking refines the list by prioritizing trials based on criterion-level relevance. A fine-tuned re-ranking LLM [38] assesses the relevance of candidate trials' eligibility criteria to the patient on a criterion-by-criterion basis and aggregates the criterion-level scores to trial-level scores, ultimately re-ranking the trials from the most to least applicable to the patient's profile. Finally, precise eligibility classification is performed using a fine-tuned Chain-of-Thought (CoT) reasoning model [33, 39], which classifies inclusion and exclusion criteria and generates a justification for each decision. This structured pipeline (see Figure 1) optimizes patient-trial matching while maintaining explainability in decision-making, and consists of four key levels:

1. *Data Ingestion and Preprocessing*: Both structured trial metadata (e.g., clinicaltrials.gov XML files) and patient records are processed, and terminology is standardized through Named Entity Recognition (NER) and entity normalization [40–44] followed by vector embedding to enable semantic search [45].
2. *Candidate Trial Retrieval*: For a given patient, a broad pool of relevant trials is obtained using a combination of BM25 lexical search with k-NN vector search (semantic retrieval) via Elasticsearch [35, 36, 46].
3. *Re-Ranking for Criterion-Level Relevance*: Trial prioritization is performed by a fine-tuned Gemma-2-2B model based on the applicability and relevance of eligibility criteria for the patient's profile.
4. *Eligibility Classification and Final Ranking*: A fine-tuned Phi-4 model performs criterion-level classification using medical Chain-of-Thought (CoT) [33, 39] reasoning, generating explanations for inclusion and exclusion decisions. A final scoring step prioritizes trials that best satisfy inclusion criteria while minimizing exclusion violations.

Table 1 provides an overview of the core components and their roles in the system; further details on model fine-tuning, candidate retrieval mechanisms, re-ranking, and criterion-level eligibility assessment are provided in the Materials and Methods section, as well as in the Supplementary Materials sections C, D, and E.

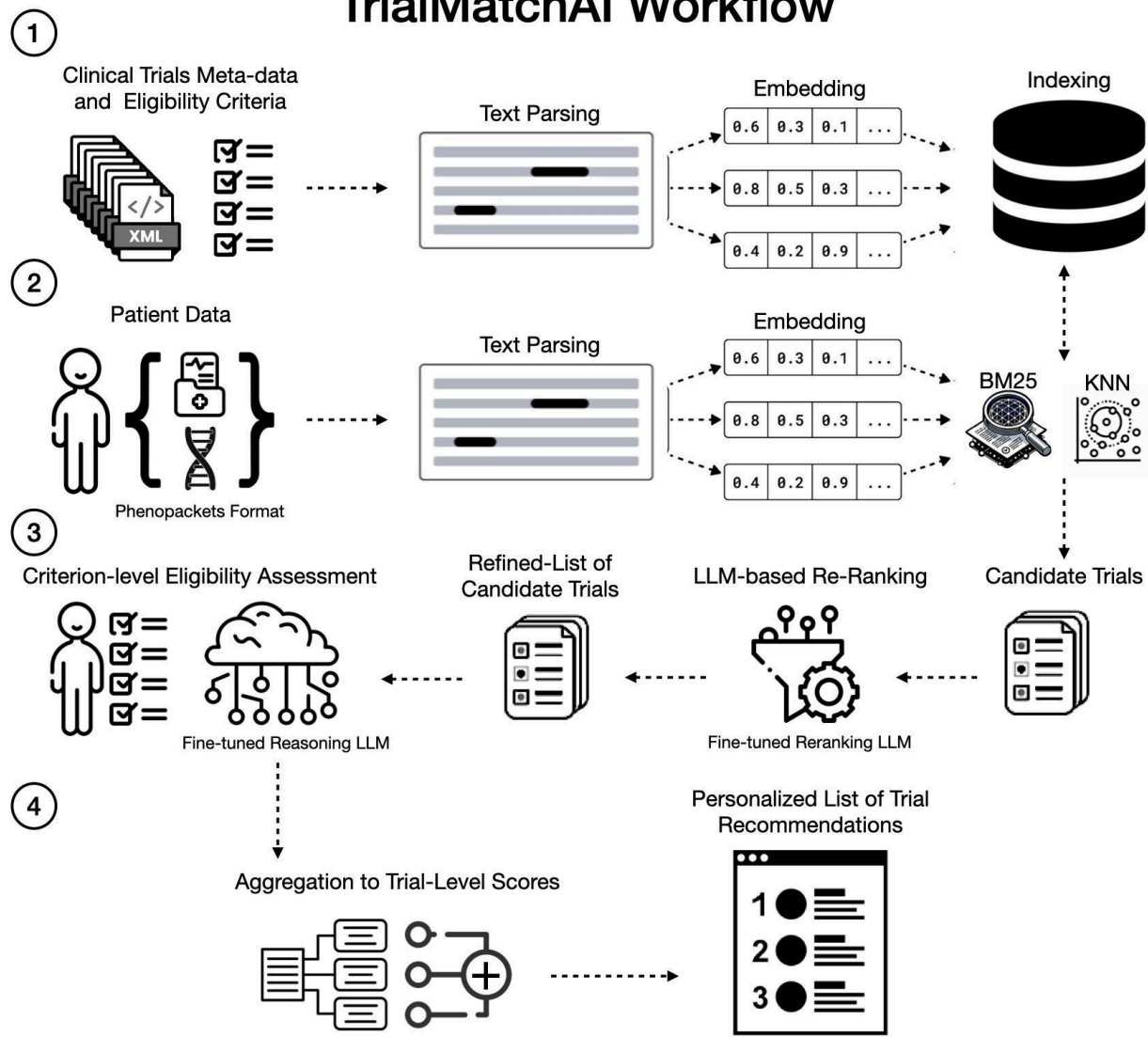

**Figure 1.** TrialMatchAI Workflow for Automated Patient-to-Trial Matching. (1) Clinical trial metadata, including eligibility criteria, are extracted from structured sources (XML formats) and processed through a text parsing module that includes named entity recognition, entity normalization, and synonym enrichment. The parsed and original texts are embedded into numerical representations using an embedding model (e.g., BGE-M3) and indexed using Elasticsearch for efficient retrieval. (2) Patient data, encompassing clinical and molecular information in Phenopackets exchange format (JSON), undergo a similar text parsing process, including named entity recognition, entity normalization, query expansion, and synonym enrichment, followed by embedding transformation to generate query vectors. (3) An initial list of candidate trials is retrieved using a hybrid approach that combines BM25 text-based retrieval with k-nearest neighbors (KNN) search on embedded representations. The list is further refined via criterion-level relevance assessment using a large language model (LLM) fine-tuned for biomedical text re-ranking (Gemma-2-2B). The relevant trials are subsequently assessed for criterion-by-criterion eligibility using a final LLM (Phi-4) fine-tuned for biomedical reasoning. The model generates concise and interpretable explanations for each criterion-level classification, ensuring the explainability of AI-driven decision-making. (4) Finally, criterion-level matching scores are aggregated into trial-level overall eligibility scores, producing a personalized ranked list of clinical trial recommendations for the given patient.

**Table 1. Summary of TrialMatchAI Components**. Asterisk (*) denotes a fine-tuned version of the model.

| Component | Role(s) | Models/Tools Used |
|---|---|---|
| **Entity Recognition** | Extract entities from unstructured text | BioBERT*, RoBERTa-large*, GLiNER |
| **Entity Normalization** | Normalize entities to knowledge bases, enrich data with synonyms | Rules-based methods, BioSyn |
| **Text Embedding** | Convert text into numerical representations | BGE-M3 |
| **Candidate Retrieval** | Perform first-stage text and semantic-based searches | Elasticsearch BM25 / KNN |
| **Neural Re-Ranking** | Refines candidate trials relevance at the criterion level | Gemma-2-2B* |
| **Recommendation Engine** | Criterion-level eligibility assessment, keyword and text generation for trial recommendations | Phi-4* |

## 2.2 Benchmarking TrialMatchAI: Synthetic Patient Evaluation

To evaluate TrialMatchAI, we utilized a comprehensive dataset comprising both in-house and publicly available resources. The in-house dataset ("Ideal Candidates" dataset) consists of synthetic patient profiles generated from a randomly sampled set of cancer-related clinical trials, specifically focusing on trials with long and complex eligibility criteria. The publicly available dataset includes synthetic summaries from the Clinical Trials (CT) tracks of the Text Retrieval Conference (TREC) [30, 31] for the years 2021 and 2022, both widely used benchmarks for assessing patient-trial matching systems [10, 30, 31]. Additionally, an expert evaluation was conducted to assess TrialMatchAI's criterion-level eligibility classification and its generated explanations. Performance was measured using multiple metrics, including accuracy, recall, normalized discounted cumulative gain (nDCG) at top-k (ndcg@5, ndcg@10, ndcg@20), and precision at top-k (p@5, p@10, p@20). See Supplementary Materials section E for metrics definitions.

**Validating AI-based Matching: The "Ideal Candidates" Dataset**

The "Ideal Candidates" dataset (see Materials and Methods, Section 4.1) serves as an initial proof of concept, demonstrating TrialMatchAI's ability to accurately align patients with their optimal clinical trial, assuming a perfect match exists. Figure 2 illustrates the distribution of ground truth trial rankings for the 100 ideal candidates. The vast majority (95%) had their assigned clinical trial ranked within the top two matches, with 92 out of 100 patients having their ground truth trial among the highest-ranked recommendations. A small subset (5%) had their ground truth trial ranked lower; however, no trial fell beyond the 9th position. This ranking discrepancy likely stems from the eligibility of some synthetic patients for multiple highly relevant trials beyond their designated ground truth trial. In such cases, TrialMatchAI identified additional trials that were equally or even more suitable based on patient characteristics, leading to slight variations in rankings. Rather than signaling a failure, this outcome highlights TrialMatchAI's potential to surface not only the expected match, but also a broader set of viable trial options—an ability that could be valuable in real-world clinical decision-making. The ground truth trial assignments and the top 10 ranked trials for each ideal candidate are provided in Supplementary Data Sheet 3.

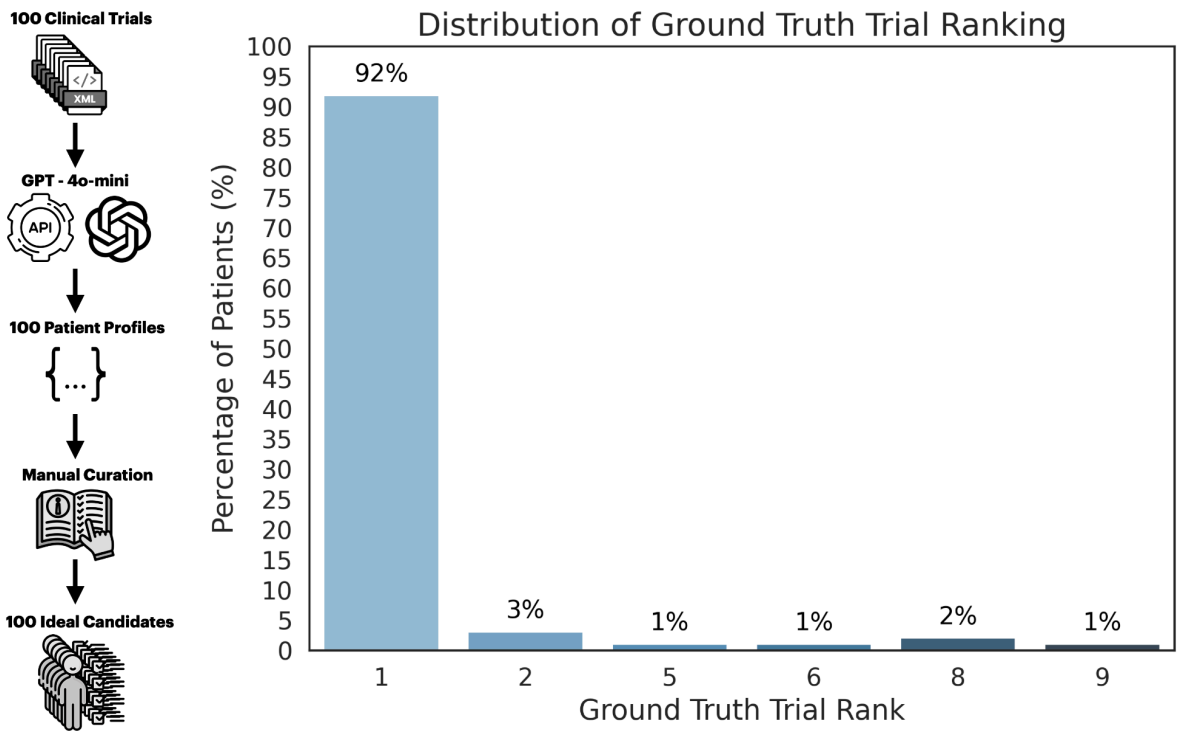

**Figure 2. Distribution of ground truth trial rankings assigned to 100 patient profiles by TrialMatchAI.** The x-axis represents the rank position of the ground truth clinical trial for each patient, while the y-axis indicates the percentage of patients assigned each rank. The vast majority of patients had their ground truth trial ranked within the top two, demonstrating high precision in patient-trial matching. The workflow on the left outlines the process: 100 clinical trials were randomly sampled, their metadata (including eligibility criteria) was presented to GPT-4o-mini via the API to generate patient profiles and manual curation was performed to obtain the final set of 100 ideal candidate profiles.

**High Recall and Precision with Accurate Trial Prioritization in TREC Benchmarks**

To assess TrialMatchAI's robustness in handling diverse and complex patient cases, we evaluated its performance on the TREC 2021 and 2022 Clinical Trials (CT) datasets [30, 31]. These datasets contain synthetic patient case descriptions resembling those found in admission notes. TREC2021 consists of 75 patient cases, while TREC2022 includes 50 patient cases. Each patient is associated with a list of trials labeled as "*Irrelevant*" (unrelated to the patient), "*Excluded*" / "*Ineligible*" (patient has the condition, but does not qualify), or "*Eligible*" (patient qualifies for enrollment). Table 3 (Materials and Methods section) provides summary statistics of the datasets.

TrialMatchAI's hybrid retrieval approach efficiently retrieves an initial subset of candidate trials from over 26,000 in each TREC dataset, reducing the search space for ranking and final eligibility assessment by more than 95% and achieving over 90% recall at just 3% of the total documents for both TREC2021 and TREC2022. This means that the majority of relevant trials for most patients are retrieved early, reducing the risk of missing eligible matches.

Figure 3A presents the recall performance of TrialMatchAI across varying retrieval sizes on the TREC datasets. Recall improves with an increasing number of retrieved trials, stabilizing around 90% at approximately 500 retrieved trials. This suggests that a cut-off at this retrieval size balances high recall with computational efficiency. The consistent recall performance across datasets highlights the model's robustness in generalizing across different patient cohorts.

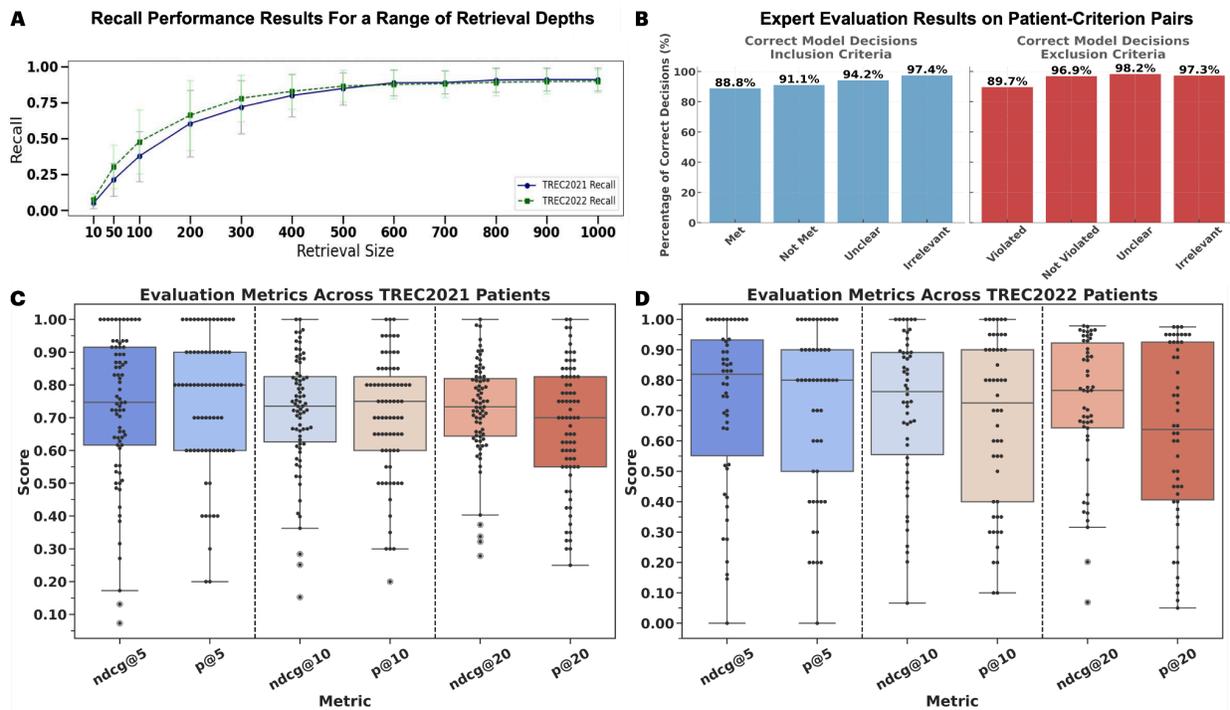

**Figure 3. Performance evaluation of TrialMatchAI on the TREC clinical trial datasets**. (A) Recall vs. retrieval size for patient-trial matching on TREC2021 and TREC2022 datasets, showing increasing recall with larger retrieval sizes. (B) Expert evaluation results on 950 patient-criterion pairs, measuring the percentage of correct model decisions for inclusion and exclusion criteria. (C) Evaluation metrics (nDCG and precision at 5, 10, and 20) across TREC2021 patients, illustrating ranking performance. (D) Evaluation metrics across TREC2022 patients, demonstrating consistent model performance. These results validate the effectiveness of the system in retrieving relevant trials and correctly assessing eligibility criteria.

Figures 3C and 3D illustrate TrialMatchAI's ranking performance on the TREC2021 and TREC2022 datasets. Ranking quality is evaluated using Normalized Discounted Cumulative Gain (nDCG) and Precision (P) at 5, 10, and 20 trials. Across both TREC datasets, TrialMatchAI achieves a median nDCG@5 of 0.74 (TREC2021) and 0.82 (TREC2022) and a median nDCG@10 and nDCG@20 of 0.75 (both), meaning the system consistently ranks the most relevant trials closer to the top. Similarly, median precision scores remained relatively high across different cut-off points. The tool achieves a median p@5 of 0.8 (for both TREC 2021 and 2022), median p@10 of 0.77 (TREC2021) and 0.72 (TREC2022), and median p@20 of 0.7 (TREC2021) and 0.62 (TREC2022). The consistency between TREC2021 and TREC2022 results underscores the model's robustness across different patient distributions.

While the metrics reported by other LLM-based solutions, notably TrialGPT [10], differ slightly in how they are aggregated, benchmark comparisons indicate that TrialMatchAI achieves highly competitive performance using significantly smaller, open-source models. TrialGPT's highest aggregated average nDCG@10 is 0.7275 and p@10 is 0.6688 across all its evaluated datasets [10]. In comparison, TrialMatchAI achieves an average nDCG@10 of 0.7232 and a higher average p@10 of 0.6865 across the TREC 2021 and 2022 benchmarks (with even stronger performance reflected in median values, see hereinabove). Additionally, TrialMatchAI outperforms all previously top-ranked systems from the official TREC 2021 and 2022 challenges [30, 31]. The top ranking model in TREC 2021, TDMINER, achieved a mean nDCG@10 of 0.715 and a mean p@10 of 0.5760, whereas the top ranking model in TREC 2022, h2oloo, achieved a mean nDCG@10 of 0.6125 and a mean p@10 of 0.5080.

## 2.3 TrialMatchAI Successfully Matches Real Cancer Patients to Biomarker-Driven Trials

Unlike general clinical trial matching, this evaluation focuses on biomarker-driven oncology trial selection, where cancer patient eligibility depends on clinical characteristics and the identified biomarkers. We incorporate real-world data from a metastatic cancer cohort derived from the Whole Genome Sequencing Implementation in Standard Diagnostics for Every Cancer Patient study (WIDE) conducted by the Netherlands Cancer Institute (NKI) [32]. See Figure 4 for baseline statistics of the selected sub-cohort of 52 patients, which had a mean age of 63.6 years (±10.3) and a sex distribution of 46.2% male and 53.8% female.

**Trial Retrieval and Ranking Performance: High Recall and Prioritization Accuracy**

Table 2 presents the evaluation results, demonstrating TrialMatchAI's effectiveness in retrieving and ranking relevant clinical trials for cancer patients. Overall Recall, representing the percentage of patients for whom at least one relevant trial was identified, increased from 84.6% within the top five recommendations (Top-5) to 92.3% within the top twenty (Top-20). This demonstrates TrialMatchAI's ability to consistently retrieve relevant trials while ensuring that the number of suggested trials remains clinically practical for review by oncologists.

Beyond recall, we assessed how well TrialMatchAI prioritizes relevant trials. The Mean Reciprocal Rank (MRR) measures how highly the first correct trial is ranked, with higher values indicating that relevant trials appear near the top of the recommendation list. The Mean Average Rank (MAR) reflects the typical rank position of relevant trials across all retrieved trials. TrialMatchAI's consistent MRR across different top-k values indicates that when a relevant trial is retrieved, it is accurately prioritized near the top of the recommendation list. This ensures that oncologists can identify the most promising trials quickly, reducing manual screening time and expediting patient enrollment.

**Table 2. Performance metrics for top-k retrieval**. Overall Recall represents the fraction of patients with at least one ground-truth trial retrieved within the top-k trials (k = 5, 10, 20). The Mean Reciprocal Rank (MRR) evaluates how well the system ranks the first ground-truth trial across all patients. MRR ranges from 0 to 1, with 1 indicating that a ground truth trial is always ranked at the top (rank 1) for every patient, and 0 indicating that no ground-truth trials were retrieved within the top-k for any patient. The Mean Average Rank is the average rank of all retrieved ground-truth trials across all patients.

| Metric | Top 5 | Top 10 | Top 20 |
| --- | --- | --- | --- |
| Overall Recall | 0.8462 | 0.8846 | 0.9231 |
| Mean Reciprocal Rank (MRR) ± Std | 0.4824 ± 0.2952 | 0.4880 ± 0.2873 | 0.4904 ± 0.2835 |
| Mean Average Rank ± Std | 2.0341 ± 0.7568 | 2.2500 ± 1.2716 | 2.8438 ± 3.1493 |

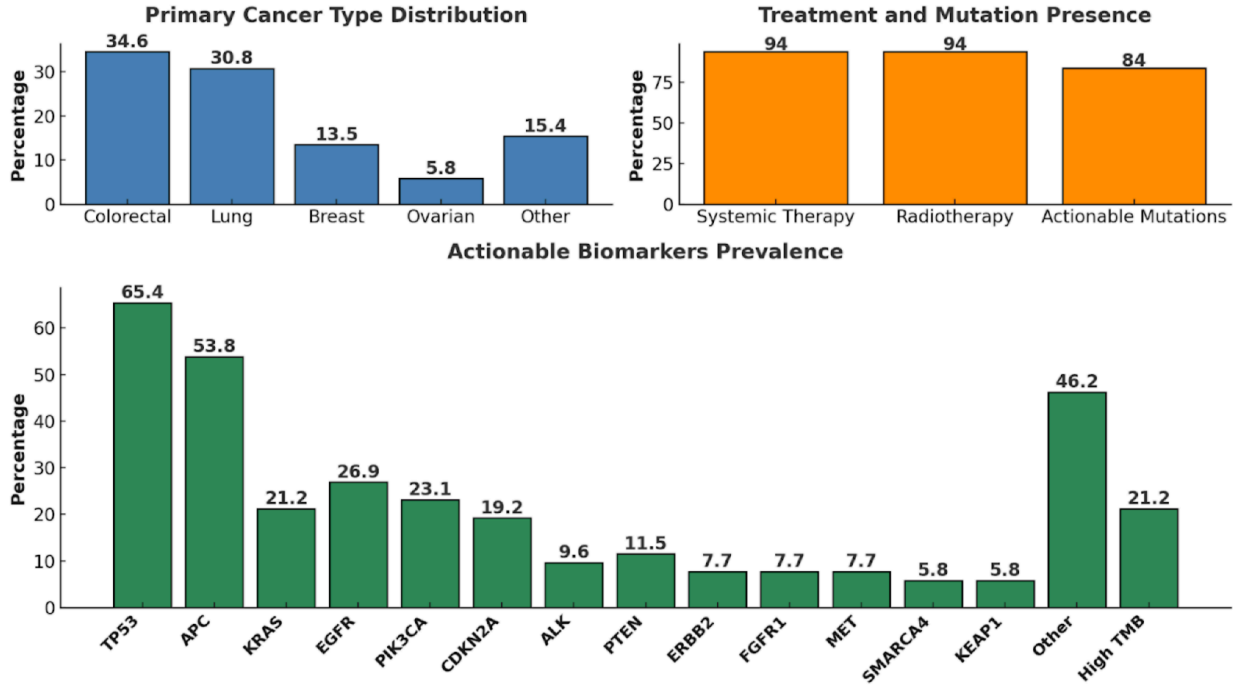

**Figure 4. Summary Statistics of the selected subset of 52 cancer patients enrolled in the WIDE study at the NKI**. The Primary Cancer Type Distribution (top-left panel) illustrates the distribution of primary cancer types in the cohort, with colorectal, lung, and breast cancer being the most prevalent. The Treatment and Mutation Presence (top-right panel) displays the proportions of patients who have undergone systemic chemotherapy and radiotherapy, along with the percentage of patients whose Molecular Tumor Board (MTB) report discusses actionable mutations. The Actionable Biomarkers Prevalence (bottom panel) presents the distribution of actionable genes identified in the MTB reports, with percentages indicating the proportion of patients harboring each mutated gene. Note: 100% of patients have metastatic cancer, provided written informed consent, and have been allocated to clinical trials.

## 2.4 Expert Evaluation of Criterion-Level Eligibility Classification

To assess TrialMatchAI's accuracy in evaluating individual eligibility criteria, we conducted an expert review of 950 randomly selected patient-criterion pairs derived from synthetic patients in the TREC 2021 and 2022 datasets. The fine-tuned Phi-4 CoT reasoning model correctly classified 88.8% of inclusion criteria as *"Met"* and 91.1% as *"Not Met"*. Additionally, it accurately identified 94.2% of *"Unclear"* criteria and 97.4% of *"Irrelevant"* criteria. For exclusion criteria, the model achieved an 89.7% accuracy in detecting *"Violated"* criteria while correctly classifying 96.9% of *"Not Violated"*, 98.2% of *"Unclear,"* and 97.3% of *"Irrelevant"* cases.

Building on our evaluation with synthetic TREC data, we further assessed the system's performance in a real-world clinical setting, focusing on molecular biomarker-driven eligibility criteria in oncology. To this end, we conducted an expert review of 100 biomarker-related inclusion criteria classified as "*Met*" by the model. These criteria were drawn from the top 20 trials recommended by TrialMatchAI for patients in the WIDE cohort. To evaluate the system's ability to match molecular biomarker criteria, we constructed patient-criterion pairs. The criterion component consisted of the selected biomarker-related inclusion criteria from these top trials. The patient component was a semi-synthetic profile incorporating biomarker details derived from the WIDE cohort. Specifically, molecular biomarker data were extracted from original Molecular Tumor Board (MTB) reports, then completely rephrased into new, decontextualized statements using the Phi-4 model. This process removed all identifying information (e.g., patient ID, age, sex) while preserving evaluative utility. This approach generated patient profiles that accurately reflected real molecular characteristics while ensuring data privacy. TrialMatchAI correctly classified 91% of these biomarker-driven inclusion criteria as "*Met*", demonstrating its effectiveness in capturing biomolecular eligibility constraints. Full results of criterion-level expert evaluations are provided in Figure 3B and Supplementary Data Sheets 1 and 2.

# 3 Discussion

Clinical trial recruitment remains a major bottleneck in the development of new therapies. Patients, particularly those with rare or complex cancers, often struggle to find suitable trials. To address this, we developed TrialMatchAI, an open-source, modular system leveraging fine-tuned open-source LLMs for precise patient-trial matching. Our evaluations demonstrate state-of-the-art performance, comparable to systems that rely on GPT-based proprietary models such as TrialGPT [10], while addressing their limitations in terms of cost, transparency, and privacy. Notably, TrialMatchAI achieves this level of performance using lightweight, open-source models, a significant advancement given the computational demands and expense typically associated with high-performance LLMs. Unlike API-dependent systems, which require external data transfer, TrialMatchAI is designed for secure operation via local deployment within hospital infrastructures. Its modular, open-source architecture makes it a strong candidate for real-world clinical integration, offering a lightweight and privacy-preserving alternative to proprietary systems for patient-trial matching in precision oncology.

TrialMatchAI's architecture is designed for scalability, explainability, and interoperability, leveraging the Phenopackets format and local deployment to ensure regulatory compliance with the European Health Data Space (EHDS), GDPR, and HIPAA [28]. Expert validation on over 1,000 patient-criterion pairs, including molecularly-driven cases, confirmed an average accuracy exceeding 90% in eligibility assessments, demonstrating the system's robust AI-driven reasoning. The model also achieves over 90% recall within the top 3% of trials in TREC datasets, effectively prioritizing relevant options while filtering out irrelevant ones. A hybrid retrieval approach, integrating LLM-based re-ranking and medical Chain-of-Thought (CoT) reasoning, ensures accurate candidate selection, with nDCG and precision at top-k metrics indicating state-of-the-art performance. Finally, a real-world evaluation on biomarker-driven data from 52 metastatic cancer patients of the WIDE study demonstrated that 92% of patients had relevant trials successfully retrieved within the top 20 results (Table 2), validating TrialMatchAI's practical utility. The top 20 threshold is clinically meaningful, aligning with the oncologists' workflows, where reviewing a manageable number of trials is crucial for informed decision-making.

Despite integrating a fine-tuned CoT model within a RAG framework, TrialMatchAI, like all systems based on LLMs, can be susceptible to confabulations (i.e., hallucinations) and misclassifications. However, based on the expert evaluation of criterion-level classifications and explanations, confabulation occurred in less than 1% (9 out of 950 pairs, see Supplementary Data Sheet 2) cases, where the reasoning model generated an explanation that did not correspond to the provided patient's information. To address this, future work should focus on developing robust flagging and monitoring mechanisms that allow clinicians to review AI-generated recommendations and report misclassifications, enabling continuous system improvement. Additionally, integrating an Agentic Workflows architecture [47]—where AI agents dynamically collaborate to verify and refine outputs—could provide an extra layer of oversight. This approach may help reduce hallucinations, offer more contextual information to the reasoning model, and ultimately enhance trustworthiness. Another challenge is balancing computational efficiency with model size, as proprietary GPT-based models still outperform open-source alternatives in inference speed. One potential solution is knowledge distillation [48, 49], where smaller, optimized models are trained to retain the accuracy of larger models while improving computational efficiency. Finally, incomplete patient records can impact accuracy. To mitigate this, collaborative filtering techniques [50]—which leverage insights from patients with similar clinical and molecular profiles—could help fill gaps in missing data and improve recommendation reliability.

Overall, TrialMatchAI has the potential to significantly advance AI-driven personalized medicine by streamlining patient recruitment to clinical trials. By combining high accuracy, interpretability, and privacy-preserving local deployment, TrialMatchAI sets a new standard for AI-driven clinical trial matching, enabling real-world clinical adoption within secure hospital environments and research institutions.

# 4 Materials and Methods

## 4.1 Materials

**Clinical Trials Data**

To support patient-trial matching, we utilized publicly available clinical trial metadata sourced from clinicaltrials.gov, the largest registry of clinical trials. This dataset includes eligibility criteria, condition(s), study design, and intervention details. Our database included over 100,000 clinical trials spanning from 1998 to 2024, ensuring a comprehensive and up-to-date resource for matching patients with relevant trials. Approximately 60% (61,731) of these trials are interventional and cancer-related, covering a diverse range of cancer types. To identify cancer-relevant trials, we mapped the condition terms from clinicaltrials.gov records to a controlled vocabulary of cancer phenotypes. This vocabulary was constructed using the OncoTree ontology (v2021_02) [51] as a starting point and expanded with additional terms from the phenOncoX project (https://github.com/sigven/phenOncoX). Trials were classified as cancer-related if they included at least one condition term matching this curated list.. Additionally, a small subset of the cancer-related trials (*n = 136*) was obtained solely from the Dutch public repository Centrale Commissie Mensgebonden Onderzoek (CCMO) [52], which provides information on medical research involving human subjects in the Netherlands.

For TREC-based evaluation, we utilized 26,149 trials for TREC 2021 dataset and 26,581 trials for TREC 2022 dataset (see Table 3). For the 'Ideal Candidates' evaluation analysis, the complete trials corpus included 61,731 cancer-related trials. Finally, for the real cancer cohort the search space comprised a total of 217 molecularly-driven clinical trials from the Netherlands Cancer Institute (NKI). Supplementary Figures F1 and F2 (Supplementary Materials, section F) provide descriptive statistics of the curated and parsed cancer-related clinical trials in our dataset, including distribution of cancer types, trial phases, intervention types, overall status, and geographical locations.

**Synthetic "Ideal Candidates" Dataset**

In order to assess the model's ability to uncover relevant trials assuming that a perfect match exists for a given trial, we generated the "Ideal Candidates" dataset. It comprises 100 synthetic cancer patient profiles, each generated using the GPT-4o-mini model [53] and manually curated for accuracy. Each patient was generated from a randomly selected clinical trial from our trial database based on specific criteria: the trial had to be cancer-related, interventional, fairly recent (starting after 2014), and include an eligibility criteria section exceeding 500 words to ensure sufficient complexity and comprehensiveness. These selected trials were then processed via an API request to GPT-4o-mini (see Supplementary Materials section C for the prompt that was used), which was instructed to generate a perfectly matching patient profile for each trial—fully satisfying all inclusion criteria while explicitly avoiding any exclusion criteria. Check the "Code and Data Availability" section for information on accessing the supplementary data dedicated for the generated patient summaries and the patient-trial matching results.

**Synthetic TREC Clinical Trial Benchmark Datasets**

The second synthetic dataset consists of patient summaries from the TREC 2021 and 2022 Clinical Trials (CT) tracks [30, 31], which include 75 and 50 cases, respectively. These summaries mimic real-world admission notes. For each patient, clinical trials in these datasets are categorized into three groups depending on the eligibility of the patient to this trial: "*Irrelevant*" (no connection to the trial), "*Excluded*" / "*Ineligible*" (has the targeted condition but fails exclusion criteria), and "*Eligible*" (meets all enrollment requirements). Summary statistics of the TREC datasets are provided in Table 3.

**Table 3. Summary of cohort characteristics for TREC 2021 CT and TREC 2022 CT.** The reported baseline statistics comprise the following: total number of patients (N); average age along with standard deviation (in years); sex distribution (ratio of male to female); average length of patient notes (measured in words); mean and standard deviation of eligible trials per patient; mean and standard deviation of excluded trials per patient; average number of irrelevant trials per patient; and the total number of initial trials within the search space, as classified by TREC judges.

| Metric | TREC 2021 CT | TREC 2022 CT |
| --- | --- | --- |
| N (Number of patients) | 75 | 50 |
| Age (years, mean ± std) | 41.6 ± 19.4 | 35.3 ± 20.2 |
| Sex (Male/Female) | 38/37 | 28/22 |
| Topic length (words) | 156.2 ± 45.4 | 109.9 ± 21.6 |
| Eligible trials per patient | 74.3 ± 49.0 | 78.8 ± 67.3 |
| Excluded trials per patient | 80.3 ± 60.3 | 60.7 ± 65.5 |
| Irrelevant trials per patient | 323.2 ± 93.2 | 568.4 ± 164.1 |
| Initial trials | 26,149 | 26,581 |

These datasets allow for rigorous evaluation using standard ranking metrics such as Normalized Discounted Cumulative Gain (nDCG@k), Precision@k, and Recall@k (see Supplementary Materials Section E for evaluation metrics definitions). Table 3 summarizes their characteristics, including age distributions, note lengths, and trial eligibility distributions.

**Real-World Patient Data from the Netherlands Cancer Institute (NKI)**

Building on strong benchmark results (see Results section for synthetic patients), we further validated TrialMatchAI's practical utility using real-world clinical data. For real-world validation, we used data from 52 metastatic cancer patients selected from the Whole Genome Sequencing Implementation in Standard Diagnostics for Every Cancer Patient (WIDE) study [32] conducted at the Netherlands Cancer Institute (NKI). The WIDE study collected essential patient information, including age, biological sex, informed consent status, biopsy feasibility, cancer type, both primary and metastatic tumor sites, and whether the patient underwent chemotherapy and radiotherapy. Notably, 94% of patients had undergone prior systemic chemotherapy and/or radiotherapy. Importantly, the study recorded the clinical trials to which each patient had been assigned, as well as unstructured descriptions of actionable molecular biomarkers, extracted from each patient's Molecular Tumor Board report. In all, the WIDE study collected the following fields relevant for clinical trial matching:

- A de-identified unique patient ID assigned to each participant,
- Year and month of birth,
- Biological sex (M/F),
- Written informed consent (Yes/No),
- Biopsy feasibility—whether a histological biopsy can be safely obtained during routine diagnostic procedures (Yes/No),
- Presence or suspicion of metastatic disease from solid tumors (Yes/No),
- Prior systemic chemotherapy (Yes/No)
- Prior radiation therapy (Yes/No),
- Tumor type classification (Text),
- Primary and metastatic tumor locations (Text),
- IDs of clinical trials to which the patient has been oriented,
- The molecular tumor board conclusion report containing actionable mutations and potential beneficial treatments (Text).

While the accessible WIDE dataset provides key attributes for 947 patients, it lacks many detailed clinical parameters essential for precise trial eligibility assessment, including functional imaging, organ function metrics, and comprehensive comorbidity records. Thus, the global dataset presented limited overlap between the fields recorded in this study and the comprehensive information in electronic health records, which were used for the original orientation of patients to their respective trials. Moreover, some patient-trial pairs lacked any matching criteria beyond the main condition and basic demographics, limiting their utility for evaluation.

To ensure meaningful evaluation, we selected a sub-cohort of 52 patients from the WIDE study whose information in the provided study fields had a sufficient intersection with the eligibility criteria of their assigned ground-truth trials, as determined by the Molecular Tumor Board based on full EHR data. Specifically, for a patient-trial pair to be included in the selected sub-cohort, at least 75% of the available patient information in the WIDE dataset had to correspond to criteria listed in the ground-truth trials. See Figure 4 in the Results section for baseline statistics of the sub-cohort of 52 patients.

Within this selected dataset, each patient was assigned at least one ground-truth trial, with some having up to three trials they were considered for. 217 clinical trials were included in this evaluation, and trial retrieval was conducted from this predefined pool. The actionable biomarkers identified through Molecular Tumor Board (MTB) reports were a key component of the data, making this dataset particularly relevant for biomarker-driven precision oncology trials.

## 4.2 Data Ingestion and Preprocessing

To facilitate patient-trial matching, TrialMatchAI processes both structured and unstructured data from diverse sources, converting this heterogeneous information into a standardized format optimized for efficient retrieval and accurate ranking, ensuring compatibility across varied clinical data types. At the end of this preprocessing pipeline, both patient records and clinical trial eligibility criteria are converted into dense embeddings, ensuring semantic compatibility for downstream retrieval and ranking. TrialMatchAI assumes patient data is provided in the Phenopackets exchange format [29], a standardized schema that facilitates semantic interoperability and consistent data representation. Users are expected to convert data from their native formats (e.g., OMOP [54]) into Phenopackets to enable compatibility with the system. See Supplementary Materials section G for additional details on Phenopackets and an sample Phenopackets template (in JSON format).

**Clinical Trial Data Preprocessing**

Clinical trial data undergoes preprocessing before downstream language processing. Data from ClinicalTrials.gov and CCMO is originally provided in XML format, from which key fields are extracted, including:

- NCT ID, brief title, official title
- Summary and detailed descriptions
- Study start/end dates, trial locations
- Minimum/maximum age, sex
- Eligibility criteria (most critical for patient-trial matching)

Each field undergoes standard preprocessing, including lowercasing, whitespace normalization, special character removal, and punctuation cleaning. Eligibility criteria require additional processing due to their importance in clinical trial matching. These undergo text splitting to extract individual inclusion and exclusion criteria, following a methodology adapted from the European Clinical Research Infrastructure Network (ECRIN) Clinical Research Metadata Repository [55] (detailed in Supplementary Materials section A).

**Entity Recognition, Normalization, and Data Augmentation**

TrialMatchAI applies a two-step process to extract and standardize biomedical entities from both patient data and clinical trial records:

1. Named Entity Recognition (NER) extracts molecular biomarkers (genes, proteins, mutations), diseases, drugs, procedures, diagnostic tests, and clinical signs. It uses fine-tuned BERT-based models: BioBERT [22], RoBERTa-large-PM-M3-Voc [42], and GliNER [43] (zero-shot NER).
2. Entity Normalization maps extracted entities to biomedical vocabularies: MeSH, OMIM, ChEBI, Cell Ontology, NCBI Gene, and UMLS [56–61]. It uses BioSyn [41], a BERT-based entity normalization framework optimized for synonym handling. Moreover, it ensures that correct synonyms rank among top candidates when linking, improving entity disambiguation.

Additionally, for patient data, the original Phi-4 model [39] is used for zero-shot data augmentation (i.e, query expansion) by generating rephrasings, synonyms, and keyword expansions. Notably, a fine-tuned version of this model is primarily employed for eligibility classification (see Section 4.4). This augmentation enhances query versatility, enabling a broader yet precise search space for relevant clinical trials. Details on the prompt used for data augmentation are provided in the Supplementary Materials section C.

**Embedding and Text Vectorization**

To enable semantic search, TrialMatchAI converts textual data into high-dimensional dense embeddings, facilitating similarity-based retrieval. For clinical trials, eligibility criteria are embedded using BGE-M3 [45], a biomedical transformer model optimized for medical text understanding. For patient records, structured and unstructured patient descriptions in Phenopackets format are embedded using the same model to ensure alignment with trial eligibility criteria. BGE-M3 transforms clinical data into dense vector representations, enabling efficient similarity-based retrieval. The BGE-M3 model has shown remarkable performance on multiple benchmarks and supports long-form text inputs up to 8,102 tokens.

## 4.3 Candidate Trial Retrieval

To efficiently retrieve relevant clinical trials for a given patient, TrialMatchAI employs a hybrid search strategy that integrates both lexical and semantic retrieval. This two-pronged approach ensures that trials are matched based on both exact textual overlap and deep semantic similarity, improving the system's ability to identify the most relevant options even when eligibility criteria are expressed in different ways. The result of this stage is an initial list of $k$ candidate clinical trials ($k = 500$ was used in our evaluations) that are passed on to the reranking stage.

**Elasticsearch Indexing**

Clinical trial metadata and eligibility criteria are indexed using **Elasticsearch**, enabling rapid retrieval and ranking of candidate trials. Each trial is indexed at two levels to optimize search performance:

1. *Trial-Level Index*: Includes structured fields such as NCT ID, title, summary, study design, locations, and overall trial status.
2. *Criterion-Level Index*: Each inclusion and exclusion criterion is stored as an independent document, linked to the corresponding trial. This fine-grained indexing strategy allows for precise ranking based on individual eligibility conditions rather than relying solely on trial-wide metadata.

To further enhance retrieval effectiveness, indexed documents are enriched with named entities and their synonyms, extracted and normalized during preprocessing. This structured representation ensures better alignment between patient profiles and trial eligibility requirements.

**Pre-Retrieval Filtering**

Before executing the hybrid search, an initial filtering step is applied to reduce computational overhead and eliminate obviously ineligible trials. This filtering considers:

- Demographic factors, such as age, sex, and geographic location.
- Trial recruitment status, excluding closed or withdrawn trials.
- Basic eligibility requirements, such as broad inclusion criteria that clearly rule out certain patient populations (e.g., pediatric vs. adult-only trials).

By removing non-matching trials early in the pipeline, this filtering process improves both retrieval efficiency and ranking precision, ensuring that only clinically relevant trials are considered in the subsequent ranking and eligibility assessment steps.

**Hybrid Retrieval Strategy**

TrialMatchAI retrieves candidate trials using a dual-ranking strategy that combines BM25 lexical search and KNN-based semantic similarity search:

- *BM25 (Lexical Search)* [35, 36]: A probabilistic ranking function that prioritizes documents containing query terms based on their frequency and importance. Optimized with parameters $k_1 = 1.2$, $b = 0.75$, BM25 provides strong keyword-based matching, making it effective for retrieving trials that explicitly mention key patient attributes.

- *K-Nearest Neighbor (KNN) Vector Search*: A semantic retrieval mechanism based on dense embeddings generated by BGE-M3, a transformer model optimized for biomedical text. TrialMatchAI implements Hierarchical Navigable Small World (HNSW) indexing [37], which allows for efficient approximate nearest-neighbor search in high-dimensional vector space.

These two retrieval methods operate in parallel via the Elasticsearch querying mechanism, ensuring that both textually relevant and conceptually similar trials are surfaced. By integrating lexical and semantic search, TrialMatchAI is able to retrieve trials that a purely keyword-based or embedding-based system might overlook.

## 4.4 LLM-based Re-Ranking of Trials and Eligibility Assessment

After candidate retrieval, TrialMatchAI utilizes a two-stage process to refine and precisely rank clinical trial recommendations:

1. *LLM-based Re-ranking*: trials are prioritized based on the relevance of their eligibility criteria to the patient's clinical profile.
2. *Criterion-level Eligibility Assessment*: a detailed, criterion-by-criterion analysis assesses the exact alignment between patient characteristics and individual trial eligibility requirements.

This structured approach ensures that recommended clinical trials are not only relevant but also precisely tailored to each patient's unique clinical profile. Below is a detailed explanation of the two-stage process.

**LLM-Based Re-Ranking of Retrieved Trials**

To refine candidate trials, a Gemma-2-2b model, fine-tuned on an augmented medical natural language inference dataset (see section 4.6), assesses the relevance of individual eligibility criteria to the patient profile. This criterion-level re-ranking, unlike initial trial retrieval, prioritizes trials with the most pertinent criteria for subsequent eligibility assessment. The process starts with a second-level hybrid search (BM25 + vector search) to retrieve relevant eligibility criteria for each patient query. These query-criterion pairs are then input into the Gemma-2-2b model, which determines if the patient query (Statement A) sufficiently addresses the clinical trial criterion (Statement B). The model outputs a binary relevance decision, converted into a 0–1 confidence score, where higher values indicate stronger relevance.

Trial-level relevance is computed by aggregating criterion scores using methods such as maximum score, mean, square-root normalization, logarithmic normalization, or a weighted combination (e.g., 70% sqrt-normalized sum, 30% max score), balancing trials with varying numbers of relevant criteria. These aggregated scores are combined with the initial trial-level retrieval score to produce a final ranking for eligibility assessment.

**Criterion-level Eligibility Assessment**

The core module of TrialMatchAI is criterion-level eligibility assessment, which serves as the final step in a Retrieval-Augmented Generation (RAG) pipeline [34]. The Phi-4 model, fine-tuned on a medical chain-of-thought (CoT) dataset [33, 62], classifies each inclusion and exclusion criterion of a given clinical trial based on the retrieved relevant information and provided patient data within its context window. For inclusion criteria, the possible classifications are "*Met*", "*Not Met*", "*Unclear*", and "*Irrelevant*", while for exclusion criteria, the model assigns one of "*Violated*", "*Not Violated*", "*Unclear*", or "*Irrelevant*". Crucially, each classification must be accompanied by a detailed justification, relying exclusively on the explicitly provided patient information and the retrieved context from the RAG pipeline. The model's final output is structured as a JSON object, where each criterion is represented as a nested field containing its original text, assigned classification, and the model's justification, with references to the retrieved knowledge that supported its decision. The complete prompt used for instructing the Phi-4 model is provided in Supplementary Materials section C.

## 4.5 Final Ranking Procedure

The final ranking stage in TrialMatchAI assigns a composite score to each clinical trial based on the alignment between the trial's eligibility criteria and the patient's information. The scoring mechanism prioritizes trials with more satisfied inclusion criteria and fewer violated exclusion criteria while treating ambiguous cases neutrally. Criteria classified as "*Unclear*" or "*Irrelevant*" are excluded from the score computation to ensure strict adherence to the available patient data. The inclusion and exclusion scores are computed as follows:

$$S_{\text{inc}} = \frac{\sum_{i=1}^{N} w_i}{N}, \quad S_{\text{exc}} = \frac{\sum_{j=1}^{M} w_j}{M}$$

where $N$ and $M$ are the total valid inclusion and exclusion criteria, respectively, and $w_i$ $w_j$ represent weighted contributions based on classification (positive for "*Met*" / "*Not Violated*", negative for "*Not Met*" / "*Violated*").

The final composite score for ranking is given by:

$$S = \frac{S_{\text{inc}} + S_{\text{exc}}}{2}$$

where higher scores indicate better trial relevance. The trials are then sorted in descending order based on $S$.

## 4.6 Fine-tuning of LLMs for Different Tasks

TrialMatchAI's core modules utilize multiple fine-tuned LLMs primarily for biomedical entity recognition, relevance-based re-ranking, and eligibility criteria evaluation.

**LLM Fine-Tuning for Biomedical Entity Recognition**

For named entity recognition (NER), BioBERT and RoBERTa-large-PM-M3-Voc were fine-tuned to identify a wide range of biomedical entities using datasets such as BC2GM [63], BC4CHEMD [64], MACCROBAT 2018 and 2020 [65], JNLPBA [66], BC5CDR [67], tmVar [68], and NCBI-disease [69]. The fine-tuning process was designed to support a multi-task NER framework, similar to Biomedical Entity Recognition and Normalization 2 (BERN2) [22]. Multi-task NER models enable the simultaneous recognition of multiple entity types in a single inference pass, significantly improving efficiency by reducing both computational overhead and memory consumption compared to single-task NER models, which require separate models for each entity type. Comprehensive evaluation results from fine-tuning on multiple benchmark datasets are provided in the Supplementary Materials section D.

**LLM Fine-Tuning for Trial Re-Ranking**

A 4-bit quantized Gemma-2-2B model [38] was fine-tuned for trial re-ranking using QLoRA (Quantized Low-Rank Adaptation) [70]. Instruction tuning, aimed at enhancing the model's ability to interpret task-specific instructions, was performed using an augmented MedNLI dataset [71]. This dataset combined original MedNLI examples with approximately 30,000 synthetic instances generated by GPT-4o-mini. Synthetic data was created by paraphrasing and expanding randomly sampled clinical trial eligibility criteria, then generating related and unrelated patient descriptions to ensure training diversity.

QLoRA fine-tuning parameters were set as follows: rank = 16, $\alpha = 32$, warmup ratio = 0.1, learning rate = $2 \times 10^{-4}$ (with scheduler), and 1 training epoch, adhering to established best practices [72]. In QLoRA, rank ($r$) determines adaptation capacity, and α scales LoRA updates. Given that relevance detection is a binary classification task, a moderate rank $r = 16$ was chosen to balance performance and computational efficiency, as higher ranks provide diminishing returns for this task's complexity. The fine-tuned Gemma-2-2b model achieves the following evaluation results on the MedNLI test set: precision = 0.8804, recall = 0.8594, F1-score = 0.8672.

**Chain-of-Thought Fine-Tuning for Eligibility Criteria Evaluation**

QLoRA was also used to fine-tune the Phi-4 model [39] with a medical chain-of-thought (CoT) training dataset for criterion-level eligibility assessment, incorporating instruction-input-output triplets. This dataset is designed to enhance an LLM's ability to engage in multi-step reasoning, allowing it to consider multiple reasoning paths before reaching a final decision [62].

Originally developed for complex Medical Question Answering (MedQA), the dataset contains over 25,000 English-language examples aimed at improving LLMs' medical reasoning capabilities. However, in this task, the focus is on criterion-level patient eligibility assessment, where the model evaluates each eligibility criterion individually rather than answering general medical questions. The objective was to equip Phi-4 with chain-of-thought reasoning rather than simply expanding its medical knowledge base. The QLoRA fine-tuning hyperparameters for this task are: rank = 32, $\alpha = 64$, warm-up ratio = 0.1, learning rate (with a scheduler) = $2 \times 10^{-4}$, and training epochs = 1. The higher rank (32) and alpha (64) were chosen for this task to accommodate the greater complexity of chain-of-thought reasoning, as multi-step logical inference requires a more expressive adaptation capacity than binary relevance classification. The fine-tuned Phi-4 model achieves the following evaluation results on the test set: precision = 0.8678, recall = 0.8805, F1-score = 0.8741.

# 5 Data and Code Availability

The synthetic datasets for the TREC 2021 and 2022 clinical trial tracks can be accessed at [2021 TREC Clinical Trials Track](#) and http://www.trec-cds.org/2022.html. All data including the fine-tuned models (low rank adapter format), "Ideal Candidates" synthetic patients dataset and their matching results, the expert-examined patient-criteria pairs, the curated and parsed clinical trial database, the results on TREC benchmarks, and the dictionaries used for entity normalization are publicly accessible via the Zenodo repository dedicated to TrialMatchAI at [10.5281/zenodo.15045515](#). Access to the real cancer patient cohort (WIDE study) is restricted and requires a formal, justified request to the Netherlands Cancer Institute.

The TrialMatchAI source code is available on GitHub at https://github.com/cbib/TrialMatchAI.

# 6 Acknowledgments

This work was supported by the European Union under the **EOSC4Cancer** project, funded by the European Research Executive Agency (REA) under the European Union's Horizon Europe program, grant agreement ID:101058427. This work also benefited from access to the computing resources of the "CALI 3" cluster. This cluster is operated and hosted by the University of Limoges. It is part of the HPC network in the Nouvelle-Aquitaine Region in France, funded by the French government and the Region.

# Supplementary Materials

## TrialMatchAI: An End-to-End AI-powered Clinical Trial Recommendation System to Streamline Patient-to-Trial Matching

**Authors:**

Majd Abdallah[1,2], Sigve Nakken[3,4,5], Mariska Bierkens[6], Johanna Galvis[1,2], Alexis Groppi[1,2], Slim Karkar[1,2], Lana Meiqari[6], Maria Alexandra Rujano[9], Steve Canham[9], Rodrigo Dienstmann[7,8], Remond Fijneman[6], Eivind Hovig[3,5], Gerrit Meijer[6], Macha Nikolski[1,2]

**Affiliations:**
[1] University of Bordeaux, CNRS, IBGC UMR 5095, 146 Rue Léo Saignat, 33000 Bordeaux, France.
[2] University of Bordeaux, Bordeaux Bioinformatics Center, 146 Rue Léo Saignat, 33000 Bordeaux, France.
[3] Department of Tumor Biology, Institute of Cancer Research, The Norwegian Radium Hospital, Oslo University Hospital, Oslo, Norway.
[4] Centre for Cancer Cell Reprogramming, Institute of Clinical Medicine, Faculty of Medicine, University of Oslo, 0379 Oslo, Norway.
[5] Department of Informatics, University of Oslo, 0316 Oslo, Norway
[6] Department of Pathology, The Netherlands Cancer Institute, Amsterdam, The Netherlands.
[7] Oncology Data Science (ODysSey) Group, Vall d'Hebron Institute of Oncology (VHIO), Barcelona, Spain.
[8] University of Vic - Central University of Catalonia, Barcelona, Spain.
[9] European Clinical Research Infrastructure Network (ECRIN), Boulevard Saint Jacques 30, 75014, Paris, France.

**Corresponding authors:** Majd Abdallah (abdallahmajd7@gmail.com), Macha Nikolski (macha.nikolski@u-bordeaux.fr)

# A: Preprocessing Clinical Trial Eligibility Criteria

Accurately structuring eligibility criteria is essential for downstream patient-trial matching. Clinical trial registries, such as ClinicalTrials.gov, present eligibility information as unstructured text, often formatted as paragraphs, bullet points, or numbered lists with significant variability. To enable fine-grained retrieval and criterion-level reasoning, we adopted a systematic segmentation and structuring approach inspired by the methodology used in the European Clinical Research Infrastructure Network's (ECRIN) Clinical Research Metadata Repository [1].

This preprocessing step creates a standardized, machine-readable dataset of individual eligibility criteria, which serves as the foundation for lexical and semantic analysis, trial ranking, and future research in AI-driven trial matching.

**Segmentation and Structuring of Eligibility Criteria**

The preprocessing pipeline follows a two-stage segmentation process to extract and standardize individual inclusion and exclusion criteria:

1. *Initial Text Segmentation and Cleaning:*
   - The eligibility criteria block is first split using regular expressions to detect list markers, such as numeric sequences ("1.", "2."), alphabetized markers ("a)", "b)"), and symbol-based markers ("•", "→").
   - Carriage returns and whitespace inconsistencies are corrected to ensure coherence.
   - Very short fragments are consolidated, and ambiguous text breaks are resolved to prevent unnecessary segmentation artifacts.
2. *Recursive Splitting and Contextual Parsing*:
   - A secondary parsing step detects nested sub-criteria, ensuring that eligibility conditions expressed in multi-level lists or embedded definitions are appropriately segmented without losing contextual dependencies.
   - Hierarchical dependencies between primary and secondary conditions are preserved using sequence numbers, indent levels, and hierarchical identifiers to maintain contextual relationships.

This pipeline ensures that eligibility criteria can be processed consistently across trials by preserving logical structures while standardizing formatting.

**Handling Formatting Ambiguities and Over-Segmentation in Eligibility Criteria**

The high variability in trial formatting poses challenges, particularly in handling irregular structures, typographical inconsistencies, and unusually complex eligibility text. Over-segmentation may occasionally occur when:

- Criteria contain nested conditions or conditional statements, leading to multiple fragmented sub-criteria.
- Formatting artifacts, such as misplaced line breaks or incomplete list markers, disrupt logical continuity.

To mitigate fragmentation effects, TrialMatchAI applies three key strategies:

1. *Query Expansion in Retrieval:* Expands the search space to capture missing contextual elements, increasing recall.
2. *Aggregation in LLM-Based Re-Ranking*: Consolidates fragmented criteria scores into comprehensive trial-level rankings.
3. *Context Window Retention in Chain-of-Thought Reasoning*: Preserves entire eligibility blocks to allow models to interpret interdependencies.

**Future Refinements**

To further optimize extraction accuracy, future versions will explore the integration of Tool-Augmented LLMs [2] for adaptive text splitting, dynamically adjusting segmentation strategies based on trial-specific structures. These models could refine parsing by detecting inconsistencies in formatting and improving eligibility extraction for large-scale clinical trial datasets.

## B: Hybrid Entity Normalization Framework in TrialMatchAI

Accurate biomedical entity normalization is essential for mapping heterogeneous clinical data—including genes, proteins, diseases, drugs, procedures, and symptoms—to standardized ontological concepts [3]. Biomedical terminology is highly inconsistent across datasets, with the same entity referenced using synonyms, abbreviations, misspellings, and varying nomenclature depending on the source [3,4]. This inconsistency creates challenges for retrieval, interoperability, knowledge extraction, and downstream applications like clinical decision support and large-scale meta-analyses.

TrialMatchAI integrates a hybrid entity normalization framework to resolve these issues, combining rule-based methods for structured mappings and neural-network-driven models for resolving complex variations. This approach ensures data consistency, improves retrieval accuracy, and enhances interoperability across biomedical systems.

**Rule-Based Normalization: GNormPlus and Sieve-Based Methods**

To ensure robust biomedical entity linking, we employ rule-based techniques that integrate dictionary-based mapping, heuristics, and contextual refinement:

- GNormPlus [5]: Used for gene and protein normalization, this system combines dictionary-based methods, abbreviation resolution, and statistical inference to accurately map gene mentions to standardized identifiers. It also resolves ambiguity by analyzing the context in which a gene name appears.
- Sieve-Based Disease Normalization [6]: This multi-layered, sequential approach progressively refines entity linking by applying:
  - Exact string matching,
  - Synonym expansion,
  - Abbreviation resolution,
  - Contextual disambiguation,
  - Hierarchical disease mapping.

Each "sieve" refines the results from the previous layer, ensuring that ambiguities are resolved step-by-step before the final entity assignment. This method significantly improves disease normalization by progressively eliminating incorrect mappings.

We use custom-built normalization scripts for drugs, cell types, procedures, symptoms, and clinical signs, leveraging regular expressions and rule-based heuristics. These scripts efficiently handle biomedical entity variability, ensuring that extracted terms align with standardized ontologies and enhancing interoperability across datasets.

**Neural Network-Based Normalization: BioSyn Models**

While rule-based methods effectively normalize structured terms, more complex entity variations—such as overlapping synonyms, ambiguous abbreviations, and polysemous biomedical terms—require deep learning-based models. To enhance entity resolution, TrialMatchAI incorporates BioSyn [7], a fine-tuned BERT-based entity normalization model that improves entity linking through:

- Synonym marginalization: Learning entity embeddings that prioritize synonyms over less relevant terms.
- Iterative candidate selection: Refining entity matches through multiple re-ranking stages to ensure the most appropriate normalization.

TrialMatchAI integrates the following BioSyn models, each fine-tuned for specific biomedical entity types:

- *biosyn-sapbert-bc5cdr-disease* (disease normalization)
- *biosyn-biobert-bc5cdr-disease* (alternative disease normalization)
- *biosyn-sapbert-bc5cdr-chemical* (chemical & drug normalization)
- *biosyn-sapbert-bc2gn* (gene normalization)

Each model has been fine-tuned to improve entity representation, ensuring high accuracy even for ambiguous, multi-term, and non-standardized biomedical entities. The BioSyn models complement rule-based approaches by capturing deep semantic relationships that traditional methods may miss.

## B.1 Gene Normalization using GNormPlus

GNormPlus is a high-performance gene normalization tool that operates via a two-step process: named entity recognition (NER) and entity normalization. TrialMatchAI utilizes only the gene normalization component to ensure high-precision gene disambiguation for clinical trial eligibility determination. This component employs a sophisticated multi-step approach, integrating species recognition, abbreviation resolution [8], composite mention simplification, and a statistical inference network to enhance disambiguation and retrieval performance [5]. The system's effectiveness has been previously validated on public benchmark datasets, including the BioCreative II Gene Normalization challenge, where it achieved an F1 score of 86.7% [5]. The following sections detail the core steps of the normalization process.

## B.2 Disease Normalization via a Sieve-Based Approach

In our system, we implemented a multi-pass sieve-based methodology for disease normalization inspired by the work of D'Souza and Ng [6]. This approach progressively refines entity normalization through a sequence of sieves (i.e., runs), each designed to address specific challenges in mapping disease mentions to standardized ontology concepts:

- *Exact Match Sieve*: This initial sieve directly maps disease mentions to ontology concepts using exact string matching. It leverages standardized vocabularies and ontologies, such as the Unified Medical Language System (UMLS), to ensure precise mappings.
- *Synonym Resolution Sieve*: This sieve utilizes curated synonym dictionaries and abbreviation expansion methods to expand normalization coverage. Recognizing various lexical forms and synonyms of disease terms, enhances the system's ability to accurately map diverse expressions of the same concept.
- *Contextual Disambiguation Sieve*: In cases where disease mentions are ambiguous, this sieve employs surrounding textual context to resolve ambiguities. It analyzes co-occurrence patterns with other biomedical entities and utilizes machine learning models trained on annotated corpora to discern the most appropriate mappings based on context.

## B.3 BioSyn for Enhanced Neural Normalization

We employ BioSyn, a neural entity linking framework that utilizes a synonym-marginalization technique to further enhance normalization accuracy. BioSyn comprises multiple fine-tuned SapBERT (Self-alignment pretraining for BERT) and BioBERT [9, 10] models to improve entity representations and disambiguation by learning robust embeddings using:

- Synonym Marginalization: Trains deep contextualized representations that incorporate all known synonyms of an entity, reducing lexical variability.
- Iterative Candidate Selection: Dynamically refines entity normalization by incorporating hard negative samples during training.

BioSyn has been shown to achieve state-of-the-art performance across multiple biomedical entity normalization benchmarks, significantly outperforming traditional string-matching and rule-based approaches [7].

# C: LLM Prompting in TrialMatchAI

## C.1 Prompting Gemma-2-2b to Evaluate the relevance of Query-Criterion Pairs

The fine-tuned Gemma-2-2b model is designed to assess the relevance of each query-criterion pair, presented as Statements A and B, using the following prompt.

```
#### System Instruction
You are a **clinical assistant** tasked with determining whether the
patient information (**Statement A**)
provides enough details to evaluate whether the patient satisfies or
violates the clinical trial eligibility criterion (**Statement B**).

- Respond with **Yes** if **Statement A** contains sufficient information
```

```
to make this evaluation.
- Respond with **No** if it does not.
---
#### Input Format
**Statement A:** {patient_text}
**Statement B:** {criterion_text}
---
#### Response Format
Your response should only be "Yes" or "No". Do not provide any additional
commentary or context.
```

## C.2 Chain-of-Thought Prompting: Criterion-level Eligibility Assessment

The fine-tuned Phi-4 model is prompted to perform medical Chain-of-Thought (CoT) reasoning to assess a patient's eligibility for a clinical trial. The prompt begins with a system priming statement, ensuring the model engages in CoT reasoning before addressing the task or question. Chain-of-thought (CoT) reasoning [11, 12] is a prompting technique designed to enhance the reasoning capabilities of large language models (LLMs). By prompting models to generate intermediate reasoning steps, CoT enables them to tackle complex, multi-step problems more effectively. This approach improves performance on arithmetic, commonsense, and symbolic reasoning tasks and offers a transparent window into the model's decision-making process. Notably, CoT prompting has demonstrated the ability to elicit strong reasoning skills in sufficiently large language models, allowing them to decompose intricate problems into structured, logical steps. Fine-tuning Phi-4 for CoT ensures that it consistently initiates its response with a structured "thinking" phase before providing the final answer, reinforcing a systematic and reliable approach to clinical trial eligibility assessment. The CoT prompt for criterion-level eligibility assessment is provided below.

```
#### Role Definition
- You are a **medical expert** with advanced knowledge in:
  - Clinical reasoning
  - Diagnostics
  - Treatment planning
- You are **detail-oriented** and think **step-by-step** to ensure
**logical and accurate** assessments.
---
### Assessment Task
- Evaluate a patient's eligibility for a clinical trial.
- Analyze each **criterion individually and systematically**.
---
### Evaluation Framework
```

```
#### 1. Inclusion Criteria Assessment
- For each inclusion criterion, classify it as:
  - **Met**: The patient's data **explicitly and unequivocally** satisfies the criterion.
  - **Not Met**: The patient's data **explicitly contradicts or fails to satisfy** the criterion.
  - **Unclear**: Insufficient or **missing data** to verify compliance.
  - **Irrelevant**: The criterion **does not apply** to the patient's context.

#### 2. Exclusion Criteria Assessment
- For each exclusion criterion, classify it as:
  - **Violated**: The patient's data **explicitly and unequivocally** violates the criterion.
  - **Not Violated**: The patient's data **confirms compliance** with the criterion.
  - **Unclear**: Insufficient or **missing data** to verify compliance.
  - **Irrelevant**: The criterion **does not apply** to the patient's context.

---

### Important Guidelines
- **Assess each criterion one by one** with a strict **evidence-based approach**.
- **Do not infer, assume, or extrapolate** beyond the **provided patient data**.
- **Justifications** must be based strictly on **direct evidence** from the patient's profile.
---
### Response Format
{
  "Inclusion_Criteria_Evaluation": [
      {
      "Criterion": "Exact inclusion criterion text",
      "Classification": "Met | Not Met | Unclear | Irrelevant",
      "Justification": "Clear, evidence-based rationale using ONLY provided data"
      }
  ],
  "Exclusion_Criteria_Evaluation": [
      {
      "Criterion": "Exact exclusion criterion text",
```

```
      "Classification": "Violated | Not Violated | Unclear | Irrelevant",
      "Justification": "Clear, evidence-based rationale using ONLY provided data"
    }
  ],
  "Recap": "Concise summary of key qualifying/disqualifying factors",
  "Final Decision": "Eligible | Likely Eligible | Likely Ineligible | Ineligible"
---
### Input Structure
#### Clinical Trial Criteria
---Start of Clinical Trial Criteria---
{eligibility_criteria_text}
---End of Clinical Trial Criteria---
#### Patient Profile
---Start of Patient Description---
{patient_profile}
---End of Patient Description---
```

## C.3 Data Augmentation Prompting of Phi-4

The data augmentation prompt directs the pre-trained (not fine-tuned) Phi-4 model [13] to generate a list of primary medical conditions (at least one) relevant to a given patient's profile (e.g., summary or note), along with other related conditions or medical keywords. This expanded set of keywords is then appended to the extracted biomedical entities and their synonyms. By facilitating query expansion, this augmentation step enhances retrieval performance, improving the accuracy and relevance of downstream processes such as clinical trial retrieval, re-ranking, and eligibility assessment. The following prompt has been used for this purpose.

```
You are a helpful medical assistant specializing in clinical trial
matching. Analyze the patient's description and extract key conditions for
trial matching.

1. **Primary Conditions**:
    - Identify the most important conditions based on the overall patient
context.
    - Provide up to 10 synonyms or variations of each condition (if
applicable).
    - Store the main conditions and their synonyms in the
"main_conditions" list.
```

```
2. **Other Conditions**:
    - Generate up to 50 additional clinically relevant items
(comorbidities, molecular markers, prior treatments, medical history,
etc.).
    - Store them in the "other_conditions" list.
3. **Expanded Patient Descriptions**:
    - Using the generated keywords (primary + synonyms + other), create
semantically rich, expanded, and diversified sentences that capture the
patient's condition.
    - **Important**: The expanded statements must be direct expansions,
rephrases, or rearrangements of the original patient note. Do not introduce
new information or make inferences beyond what is explicitly stated.
    - Include the expanded sentences in an "expanded_sentences" field.
**Output**:
Return only a valid JSON object (no extra text) following this structure:

```json
{
    "main_conditions": ["Condition", "Alias1", "Alias2", "..."],
    "other_conditions": ["AdditionalCondition1", "AdditionalCondition2",
"..."],
    "expanded_sentences": [
    "Original patient sentence1...",
    "Original patient sentence2...",
    "...",
    "Expanded sentence for Condition1...",
    "Expanded sentence for Condition2...",
    "..."
    ]
}
```

## C.4 Generation of "Ideal Candidates" Prompting of GPT-4o-mini

To generate 100 synthetic patient profiles as ideal candidates for 100 pseudo-randomly selected clinical trials, we use the following prompt to instruct GPT-4o-mini [14]. The prompt provides the model with the condition, age, sex, and eligibility criteria section of a selected clinical trial, guiding it to generate a complete and realistic patient profile.

```
You are a seasoned medical expert. Based on the clinical trial details
provided below, generate a concise and realistic one-paragraph patient
summary that resembles an admission or electronic health record note
```



```
describing an ideal candidate for the trial with {condition_info}, whose
age falls between {minimum_age} and {maximum_age} and whose sex is
{biological_sex}.

**Requirements:**
- The patient's medical history and current condition must strictly satisfy
**every inclusion criterion** without exception.
- The note must also ensure that the patient **clearly and explicitly does
not violate any exclusion criteria**.

**Trial Information:**
-   **Sex   and   Age   Requirements:**   {biological_sex},   {minimum_age},
{maximum_age}
- **Eligibility Details:** {eligibility_criteria_block}

**Guidelines for Writing the Note:**
- Integrate the provided information naturally into the note, maintaining
an electronic health record (EHR) or admission note style.
- Avoid any explicit reference to the trial, eligibility criteria, or the
patient's suitability for participation. Do not use phrases such as *meets
the corresponding criterion*, *satisfies the requirements of the trial*, or
similar statements.
- Ensure that all key inclusion and exclusion criteria are methodically and
seamlessly incorporated into the patient's narrative while maintaining
fluency and coherence.
```

## D: Fine-tuning of BioBERT and Roberta-large on NER Results

The fine-tuning results of BioBERT and Roberta-large-PM-M3-Voc for recognizing multiple biomedical entities are presented below. The Disease dataset consists of BIO-labeled entities from NCBI [15], MACCROBAT [16], and BC5CDR [17], encompassing a total of 14,534 labeled entities. The Gene/Protein/Mutation dataset includes 58,188 labeled entities from BC2GM [18], tmVar [19], and JNLPBA [20]. The Drug/Chemical dataset comprises 89,350 labeled entities sourced from BC4CHEMD [21], BC5CDR, and MACCROBAT. The Cell Type dataset contains 8,639 labeled entities from JNLPBA, while the DNA/RNA dataset consists of 11,658 labeled entities from the same source.

In contrast, the GliNER zero-shot model [22] was not fine-tuned on these datasets. Instead, it was prompted to recognize entities related to diagnostic tests, treatments/therapies, laboratory tests/results, surgical procedures, symptoms/signs, radiology, and genomic analysis techniques.



**Table D1** Named Entity Recognition (NER) F1 Scores of BioBERT v1.2 and Roberta-large-PM-M3-Voc on Multiple Datasets

| Entity Type | BioBERT v1.2 | | Roberta-large-PM-M3-Voc | |
|---|---|---|---|---|
| | Eval Dataset | Test Dataset | Eval Dataset | Test Dataset |
| Disease | 0.843 | 0.814 | 0.852 | 0.821 |
| Gene/Protein/Mutation | 0.754 | 0.738 | 0.801 | 0.794 |
| Drug/Chemical | 0.878 | 0.885 | 0.909 | 0.919 |
| Cell Type | 0.836 | 0.796 | 0.834 | 0.804 |
| DNA/RNA | 0.846 | 0.791 | 0.842 | 0.799 |

# E: Patient-Trial Matching Evaluation Metrics

To assess the performance of TrialMatchAI on the evaluation datasets, we employ a set of established evaluation metrics commonly used in information retrieval (IR) and ranking tasks. These metrics evaluate the system's ability to correctly identify and rank relevant clinical trials for a given patient description.

- Precision@k (P@k) measures the proportion of relevant clinical trials within the top k retrieved results. It is defined as:

$$P@k = \frac{1}{2 \times k} \sum_{i=1}^{k} rel(i)$$

where rel(*i*) is 1 if the i-th retrieved trial is relevant (i.e., the patient is eligible to join) and 0 otherwise. This metric evaluates the effectiveness of ranking models in surfacing relevant trials within the top results.

- Normalized Discounted Cumulative Gain at k (nDCG@k) evaluates ranking quality by assigning higher relevance to correctly ranked relevant trials. It is computed as follows:

$$DCG@K = \sum_{i=1}^{k} \frac{rel(i)}{log_2(i+1)}$$

$$IDCG@K = \sum_{i=1}^{k} \frac{rel_{ideal}(i)}{log_2(i+1)}$$



$$nDCG@K = \frac{DCG@K}{IDCG@K}$$

where IDCG@k represents the ideal DCG obtained by optimally ranking the relevant trials. NDCG@k is normalized between 0 and 1, where 1 indicates a perfect ranking. $rel(i)$ and $rel_{ideal}(i)$ are the predicted and ideal relevance ranking of trial $i$, respectively.

- Mean Reciprocal Rank (MRR) measures how early the first relevant clinical trial appears in the ranked list, capturing ranking effectiveness. It is computed as:

$$MRR = \frac{1}{N}\sum_{i=1}^{N}\frac{1}{rank(i)}$$

where $rank(i)$ is the position of the first relevant trial for patient $i$, and $N$ is the total number of patients. A higher MRR indicates that relevant trials appear earlier in the retrieved list.

## F: Summary Statistics of Cancer-related Clinical Trials

As of this writing, our curated clinical trial dataset includes 61,731 cancer-related trials with start dates ranging from 1998 to 2024. Over 99% of these trials are interventional and span multiple phases when applicable. **Figure F1** presents distribution plots for main cancer types (A), trial phases (B), intervention types (C), and overall trial status (D). These distributions provide a high-level summary, grouping similar categories and smoothing minor discrepancies from label variations. Note that clinical trials can target multiple cancer types or employ multiple intervention types, which can cause category percentages to sum beyond 100%. **Figure F2** shows the distribution of trial locations by country in percentage. Most cancer-related clinical trials in our database, at the time of writing, are in the United States and China.



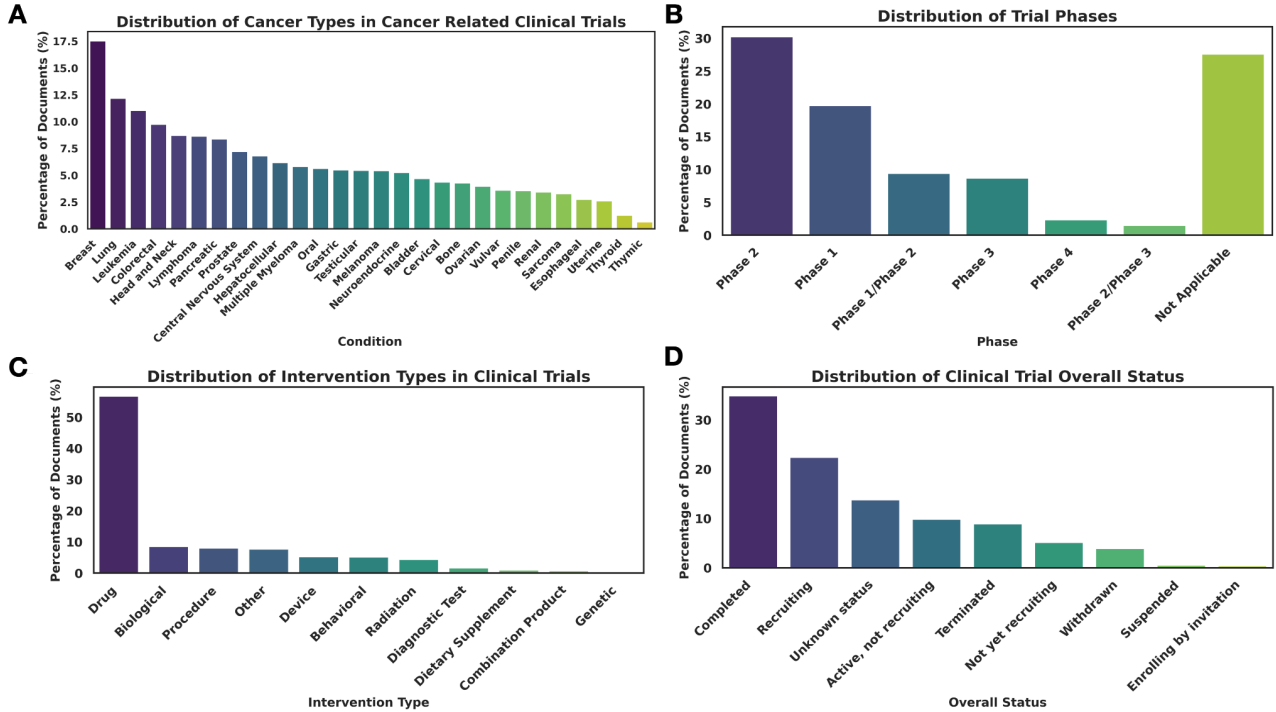

**Figure F1.** Distribution of key characteristics in cancer-related clinical trials. (A) The prevalence of different cancer types in clinical trial documents, with breast cancer being the most frequently studied. (B) The distribution of trial phases shows a higher proportion of Phase 2 and Phase 1 trials, with a significant portion categorized as "Not Applicable." (C) The distribution of intervention types highlights that drug-based interventions dominate clinical trials, followed by biological and procedural interventions. (D) The overall status of clinical trials indicates that the majority are completed, with recruiting trials being the next most common status. These distributions provide insight into the landscape of cancer clinical trials, which, at the time of writing, are utilized both for evaluating TrialMatchAI and for availability to users, with periodic updates to ensure relevance and accuracy.

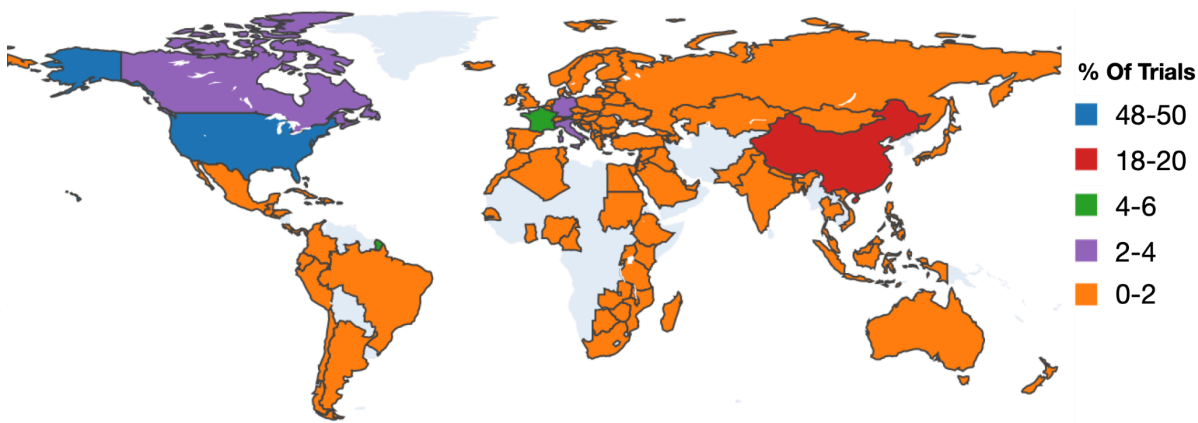

**Figure F2.** Global distribution of locations by country (%) of cancer-related clinical trials in TrialMatchAI. The map visualizes the geographic concentration of clinical trial sites, with countries shaded based on the percentage of total clinical trials conducted within their borders. The United States (blue) hosts the largest proportion of clinical trials, followed by China (red), Canada (purple), and France (purple). Many other countries, including those in Europe, South America, Africa, and parts of Asia, contribute smaller percentages (orange and green).



# G: Phenopackets Exchange Format

Patient data can be formatted in the Phenopackets exchange format [23], a structured data standard that ensures interoperability across clinical and research settings. This mitigates challenges in healthcare systems that still rely on manual entry and indexing of phenotypic data, potentially leading to inconsistencies in data exchange. Phenopackets provide a schema for encoding essential clinical and molecular attributes, including demographics such as age, sex, race, and ancestry, as well as clinical diagnoses encoded with ontologies like MONDO or ICD [24, 25]. It also includes phenotypic features mapped to the Human Phenotype Ontology (HPO) [26], genetic variants annotated with standards such as HGVS or ClinVar [27, 28], and detailed medical history, including prior treatments and interventions.

Importantly, Phenopackets supports free-text descriptions, allowing for unstructured clinical narratives alongside structured annotations, and ensures that disease progression details, symptom variability, and treatment responses are also retained, enriching the matching process. Since LLMs form the core of TrialMatchAI, both structured and unstructured textual information are incorporated into trial matching.

Below is a JSON-based Phenopackets template that can be used to provide patient data to TrialMatchAI.

```
{
  "id": "cancer-patient-example-001",
  "subject": {
      "id": "patient-001",
      "sex": "FEMALE",
      "ageAtDiagnosis": {
      "age": "P55Y"
      }
  },
  "phenotypicFeatures": [
      {"type": {"id": "HP:0001945", "label": "Fever"}},
      {"type": {"id": "HP:0002014", "label": "Weight loss"}},
      {"type": {"id": "HP:0002099", "label": "Asthenia"}},
      {"type": {"id": "HP:0002719", "label": "Anemia"}},
      {"type": {"id": "HP:0012378", "label": "Fatigue"}}
  ],
  "diseases": [
      {
      "term": {"id": "NCIT:C4872", "label": "Breast Carcinoma"},
      "clinicalTnmFinding": [
      {"value": "T2N1M0", "description": "Tumor size approximately 3 cm with regional lymph node involvement, no distant metastasis"}
      ],
      "primarySite": {"id": "UBERON:0000310", "label": "Breast"},
      "stage": {"id": "NCIT:C27971", "label": "Stage IIB"}
```




```
      }
    ],
    "biosamples": [
        {
        "id": "biosample-tumor-001",
        "sampledTissue": {"id": "UBERON:0000310", "label": "Breast"},
        "tumorProgression": {"id": "NCIT:C84509", "label": "Primary malignant neoplasm"},
        "histologicalDiagnosis": {"id": "NCIT:C4194", "label": "Infiltrating Ductal Carcinoma"},
        "procedure": {"code": {"id": "NCIT:C5189", "label": "Biopsy"}},
        "description": "Tumor biopsy from left breast showing infiltrating ductal carcinoma, moderate differentiation, ER-positive, PR-negative, HER2-positive"
        }
    ],
    "treatments": [
        {"agent": {"id": "NCIT:C405", "label": "Doxorubicin"}, "routeOfAdministration": {"id": "NCIT:C38288", "label": "Intravenous Route"}},
        {"agent": {"id": "NCIT:C1647", "label": "Trastuzumab"}, "routeOfAdministration": {"id": "NCIT:C38288", "label": "Intravenous Route"}}
    ],
    "interpretations": [
        {
        "id": "interpretation-001",
        "diagnosis": {
        "disease": {"id": "NCIT:C4872", "label": "Breast Carcinoma"},
        "genomicInterpretations": [
            {
            "status": "POSITIVE",
            "gene": {"id": "HGNC:1100", "symbol": "BRCA1"},
            "variantInterpretation": {
            "variationDescriptor": {"id": "ClinVar:17661", "label": "BRCA1 c.68_69delAG (p.Glu23Valfs)"},
            "therapeuticActionability": "Potential sensitivity to PARP inhibitors"
            }
            }
        ]
        },
        "description": "Patient diagnosed with BRCA1-positive infiltrating
```




```
ductal carcinoma, ER-positive, HER2-positive; recommended genetic
counseling, targeted therapy with trastuzumab, and consideration for PARP
inhibitors."
    }
  ],
  "metaData": {
      "created": "2024-03-24T00:00:00Z",
      "createdBy": "Your Institution Name",
      "phenopacketSchemaVersion": "2.0",
      "resources": [
      {"id": "hp", "name": "Human Phenotype Ontology", "url":
"http://purl.obolibrary.org/obo/hp.owl", "version": "2024-03-01"},
      {"id": "ncit", "name": "NCI Thesaurus", "url":
"https://ncit.nci.nih.gov/ncitbrowser/", "version": "23.02d"},
      {"id": "uberon", "name": "Uber-anatomy ontology", "url":
"http://purl.obolibrary.org/obo/uberon.owl", "version": "2024-01-15"},
      {"id": "hgnc", "name": "HUGO Gene Nomenclature Committee", "url":
"https://www.genenames.org", "version": "2024-03-01"},
      {"id": "clinvar", "name": "ClinVar", "url":
"https://www.ncbi.nlm.nih.gov/clinvar/", "version": "2024-03-01"}
      ]
  }
}
```